\documentclass{article}

\usepackage{PRIMEarxiv}

\usepackage[utf8]{inputenc} % allow utf-8 input
\usepackage[T1]{fontenc}    % use 8-bit T1 fonts
\usepackage{hyperref}       % hyperlinks
\usepackage{url}            % simple URL typesetting
\usepackage{booktabs}       % professional-quality tables
\usepackage{amsfonts}       % blackboard math symbols
\usepackage{nicefrac}       % compact symbols for 1/2, etc.
\usepackage{microtype}      % microtypography
\usepackage{lipsum}
\usepackage{fancyhdr}       % header
\usepackage{graphicx}       % graphics
\usepackage{hyperref}
\usepackage{graphicx} 
\usepackage{booktabs}
\usepackage{float}
\usepackage{algorithm}
\usepackage{algorithmic}
\usepackage{amsmath}
\usepackage{siunitx}
\usepackage{silence}
\usepackage{xcolor}
\usepackage{multirow} % 添加此行以支持 \multirow 命令
\usepackage{multicol}
\usepackage{tabularx}
\usepackage{subcaption} % 在导言区引入
\graphicspath{{media/}}     % organize your images and other figures under media/ folder

\title{LLM Hallucination Detection: A Fast Fourier Transform Method Based on Hidden Layer Temporal Signals
\thanks{\textit{\underline{Citation}}: 
\textbf{Authors. Title. Pages.... DOI:000000/11111.}} 
}

\author{
  Jinxin Li,
  Gang Tu\thanks{Corresponding Author}, 
  ShengYu Cheng,   Junjie Hu ,Jinting Wang, Rui Chen, Zhilong Zhou, Dongbo Shan\\
  School of Computer Science and Technology \\
  Huazhong University of Science and Technology \\
  Wuhan, China\\
  \texttt{\{lijinxin\_1,tugang\}@hust.edu.cn} \\
}

\begin{document}
\maketitle

\begin{abstract}
% \lipsum[1]
Hallucination remains a critical barrier for deploying large language models (LLMs) in reliability-sensitive applications. Existing detection methods largely fall into two categories: factuality checking, which is fundamentally constrained by external knowledge coverage, and static hidden-state analysis, that fails to capture deviations in reasoning dynamics. As a result, their effectiveness and robustness remain limited.
We propose HSAD (Hidden Signal Analysis-based Detection), a novel hallucination detection framework that models the temporal dynamics of hidden representations during autoregressive generation. HSAD constructs hidden-layer signals by sampling activations across layers, applies Fast Fourier Transform (FFT) to obtain frequency-domain representations, and extracts the strongest non-DC frequency component as spectral features. Furthermore, by leveraging the autoregressive nature of LLMs, HSAD identifies optimal observation points for effective and reliable detection.
Across multiple benchmarks, including TruthfulQA, HSAD achieves over 10 percentage points improvement compared to prior state-of-the-art methods. By integrating reasoning-process modeling with frequency-domain analysis, HSAD establishes a new paradigm for robust hallucination detection in LLMs.
\end{abstract}

% keywords can be removed

\section{Introduction}
Large Language Models  have demonstrated exceptional performance in tasks such as language understanding and code generation in recent years, establishing themselves as the foundational technology for various applications, including text generation, question-answering systems, and information extraction \cite{Llama_2}. However, the frequent occurrence of hallucinations during their generation process—defined as the production of results that are factually inconsistent or lack contextual support—not only undermines the credibility of LLM outputs but also severely restricts their deployment in high-stakes scenarios \cite{ju2024large, li2024dawn}.

\begin{figure*}[h]
  \centering
  \begin{minipage}[h]{0.55\textwidth}
Inspired by this, the HSAD method models the hidden layer vectors from different stages of the LLM's reasoning process as a hidden layer temporal signal, analogous to the process of human deception detection, to perform hallucination detection on the LLM.
  
 Specifically, as depicted in Figure~\ref{fig:overview}, HSAD first extracts hidden layer vectors from the LLM's forward reasoning process and constructs them into a hidden layer temporal signal with temporal characteristics, ordered by layer. Subsequently, it employs the Fast Fourier Transform  to map this temporal signal into the frequency domain, thereby constructing spectral features that reveal anomalous signals during reasoning. Experiments have demonstrated that these anomalous signals often correspond to hallucinations in the generated content. Building upon this foundation, the spectral features are further utilized as a discriminative basis to design a hallucination detection algorithm for identifying potential hallucinatory content within the LLM's generation process.
  \end{minipage}%
  \hfill
  \begin{minipage}[h]{0.4\textwidth}
    \centering
    \includegraphics[width=\linewidth]{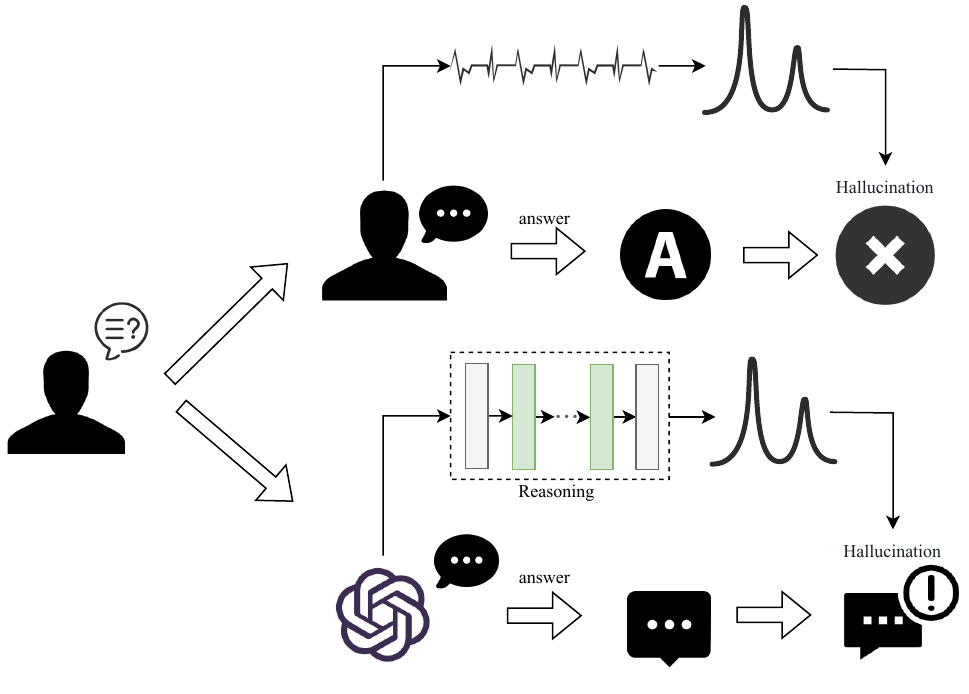}
    \caption{A comparison diagram of human lie detection vs. model’s lie detection.}
    \label{fig:overview}
  \end{minipage}
\end{figure*}

In cognitive neuroscience, a substantial body of experimental evidence indicates that when humans encounter scenarios involving information falsification or cognitive conflict, they exhibit specific psychological and neural signal changes over time. These include a gradual increase in cognitive load, fluctuations in attentional states, and the evolution of electroencephalogram  signals in their spectral structure \cite{lo2018wmdeception}. As illustrated in Figure~\ref{fig:overview}, these time-evolving signal patterns can be regarded as observable physiological manifestations of internal conflict, which indirectly reflect an individual's cognitive state and the dynamic process of information processing. The analysis of such signals can provide a theoretical basis and technical support for the study of human deception detection. It has been empirically demonstrated that LLM often exhibit behavioral patterns similar to those of humans under circumstances of information fabrication. Therefore, by drawing inspiration from the signal modeling approaches in cognitive neuroscience, the LLM's reasoning process can be modeled as a temporal signal, offering a novel perspective for hallucination detection.

This strategy aims to characterize the evolutionary trajectory of the LLM's thought process in an interpretable manner, thereby enabling the effective detection of hallucinations \cite{lieberum2023gemma, lindsey2025biology}.
Unlike most existing methods, HSAD does not rely on external knowledge bases for fact verification, but instead focuses on detecting anomalies within the LLM's internal signals. 
The main contributions of this paper are summarized as follows:

\begin{itemize}
  \item We propose an LLM hallucination detection method, HSAD, which is analogous to the human deception detection mechanism. This method constructs a hidden layer temporal signal based on the LLM's inference process and introduces frequency-domain analysis and spectral feature construction to analyze and identify potential hallucinations. Theoretical derivations and experiments have demonstrated the feasibility and effectiveness of this approach.

  \item We design and implement an LLM hallucination detection algorithm based on spectral features. Specifically, we derive and determine the observation points for LLM hallucinations, then construct spectral features in the frequency domain for detection. This algorithm achieves SOTA performance on multiple standard datasets, while ablation experiments verify the individual contributions and synergistic gains of each module.
\end{itemize}
% \lipsum[2]
% \lipsum[3]

\section{Related Work}

Existing hallucination detection methods largely build on LLM interpretability research, which can be grouped into two directions: (1) fact-consistency verification and (2) hidden representation analysis. Interpretability studies aim to uncover internal mechanisms and decision logic of LLMs, providing a foundation for improving reliability and controllability~\cite{ji2024anah}. Representative approaches include observation-based methods (e.g., probing analysis, Logit lens, sparse representations, cross-model explanations), intervention-based methods (e.g., activation patching~\cite{zhao2024automated}), hybrid methods, and other techniques such as transcoders~\cite{turpin2023language}, concept-driven explanations~\cite{wan2024acueval}, classifier injection, and SVD-based attention analysis. These methods offer theoretical support for identifying hallucination-related features and advancing systematic understanding of LLM behaviors.

\textbf{Fact-Consistency Verification.}
This line of work aligns generated content with external authoritative knowledge to detect misinformation~\cite{zhao2024fact}. FActScore~\cite{min2023factscore} segments generations into atomic facts and verifies them against knowledge bases such as Wikipedia, but suffers from limited coverage and update frequency. SAFT~\cite{rawte2024tutorial} extends this idea with search-engine interactions, where LLM agents conduct multi-round verification to enhance factual evaluation of long texts, albeit at high computational cost and with the same knowledge coverage limitations.

\textbf{Hidden Representation Analysis.}
Another direction analyzes internal representations to detect hallucinations~\cite{park2025steer}. INSIDE~\cite{chen2024inside} parses dense semantic signals and applies feature clipping to regulate activations, reducing overconfidence. Duan et al.~\cite{greenblatt2024alignment} empirically showed distinct hidden states for truthful vs. fabricated responses, while Zhou et al.~\cite{he2024llm} applied Fourier analysis to reveal frequency-domain features. Probing-based studies further explore layerwise knowledge encoding~\cite{ju2024large}, concept specialization, and syntactic learning depth~\cite{he2024decoding}. Recent methods also construct explicit latent spaces: TTPD~\cite{burger2024truth} defines a “truth subspace” robust to negation and complex expressions, and TruthX~\cite{zhang2024truthx} disentangles semantic and factual dimensions via autoencoding, editing hidden states to mitigate hallucinations.

\section{Modeling and analysis}

To understand the internal cognitive mechanisms of LLM during the text generation process, this chapter establishes a formal mathematical model based on its forward-pass reasoning process and provides a criterion for hallucination discrimination. Subsequently, from the perspective of temporal modeling, the hidden layer vectors in different layers are regarded as the temporal unfolding of the reasoning process, thereby laying the foundation for subsequent frequency-domain modeling.

\subsection{Modeling of LLM Inference Processes}

This section systematically elaborates on the computational logic and symbolic system of hidden vectors in each layer of the LLM. 

On this basis, a unified representation of the inference process is proposed and illustrated through diagrams.

\begin{figure*}[h]
  \centering
  \begin{minipage}[h]{0.48\textwidth}
    The attention vector $ah_i^j$ of token $t_i$ at layer $j$ is given by Equation~\ref{eq:1}.
    \begin{equation}
      ah_i^j = ATT(h_i^{j-1}, k_{cache}, v_{cache})
      \label{eq:1}
    \end{equation}
    
    The MLP vector $mh_i^j$ of token $t_i$ at layer $j$ is given by Equation~\ref{eq:2}.
    \begin{equation}
      mh_i^j = MLP(rh_i^j)
      \label{eq:2}
    \end{equation}
    
    % where $rh_i^j$ is the residual vector of the attention mechanism.
    
    The final hidden vector $h_i^j$ of token $t_i$ at layer $j$ is given by Equation~\ref{eq:3}.
    \begin{equation}
      h_i^j =  (ATT \circ MLP)(h_i^{j-1})
      \label{eq:3}
    \end{equation}
    
    % Each $Layer^j()$ above consists of an attention sub-layer and an MLP sub-layer, forming the main path for cross-layer information transmission.
    
    The processing of tokens $t_0,\dots,t_i$ by the LLM is defined as the $i$-th inference step $F_i$, as shown in Equation~\ref{eq:4}.
    \begin{equation}
      \begin{split}
        F_i(t_0, \dots, t_i) = {}& (emb \circ Layer^1 \circ \dots \circ Layer^l \\
                            & \circ unemb)(t_0, \dots, t_i)
      \end{split}
      \label{eq:4}
    \end{equation}
    Finally, the frequency-domain representation vector $f$ is constructed, as shown in Equation~\ref{eq:11}.
\begin{equation}
f = [A^{1}, A^{2}, \dots, A^{d}] \in \mathbb{R}^d
\label{eq:11}
\end{equation}
  \end{minipage}
  \begin{minipage}[h]{0.48\textwidth}
    \centering
    \includegraphics[width=\linewidth]{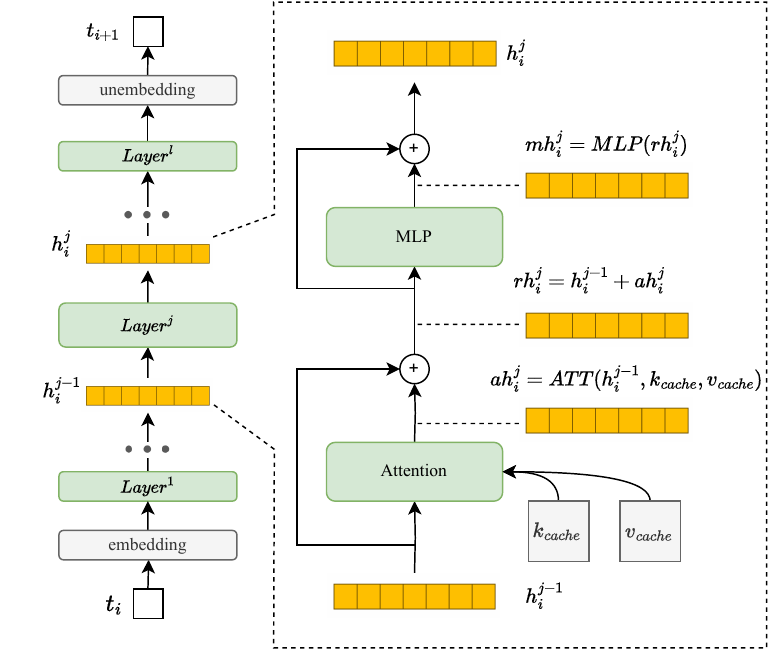}
    \caption{Schematic diagram of  $F_i$}
    \label{fig:LLM}
  \end{minipage}
\end{figure*}

Let the input question be denoted as $Q = (t_0^Q, t_1^Q, \dots, t_{m-1}^Q)$ and the corresponding reference answer as $A = (t_0^A, t_1^A, \dots, t_{n-1}^A)$, where $t_i$ represents the $i$-th token in the input or output sequence.
As illustrated in Figure~\ref{fig:LLM}, for an LLM with $l$ decoder layers, the computation at each layer can be expressed as follows.

As shown in Figure~\ref{fig:LLM}, this frequency-domain vector $f$ represents the spectral features of the input $Q$, which can effectively characterize the anomalous perturbations in LLM during inference and provide structured input for hallucination detection. 
    
  % \end{minipage}%

% As shown in Figure~\ref{fig:LLM}, 
During the LLM inference process, the hidden vectors are continuously updated and integrated with the semantic context, exhibiting distinct temporal characteristics and information flow. 
Therefore, the entire inference process can be analogized to a thought process that evolves over time. 
Based on this, the hidden vectors from different layers are constructed into hidden layer temporal signals for modeling and analysis in the frequency domain.

\subsection{LLM hidden layer temporal signal construct}

The LLM hidden layer temporal signal is defined as $X \in \mathbb{R}^{4l}$, which essentially represents a cross-layer sampling of hidden vectors, used to analyze the variation characteristics of the LLM's internal representations with increasing layer depth.

The specific construction process of the hidden layer temporal signal is as follows.

Assume the LLM has $l$ layers of Transformer architecture, with each layer containing an attention sub-layer and an MLP sub-layer, both with residual connections. From each layer, we sample the vectors of four key nodes corresponding to the generated token $t_{i+1}$, which are concatenated in the $j$-th layer to form a $d \times 4$ matrix $\mathbf{V}^j$, as shown in Equation~\ref{eq:5}.

\begin{equation}
\mathbf{V}^j = 
\begin{bmatrix}
h_{i}^j  \\
mh_{i}^j \\
rh_{i}^j \\
ah_{i}^j 
\end{bmatrix}^T \in \mathbb{R}^{d \times 4}
\label{eq:5}
\end{equation}

The four key node vectors are respectively: (1) The attention output vector $ah_{i}^j$: reflects the modeling result of the current layer's contextual dependencies; (2) The attention residual vector $rh_{i}^j$: superimposes previous layer information with the attention output; (3) The MLP output vector $mh_{i}^j$: embodies the local decision basis after nonlinear feature mapping; (4) The $j$-th layer output vector $h_{i}^j$: the fused result of all sub-module outputs in this layer.

Concatenate all $\mathbf{V}$ matrices from each layer $l$ to construct the matrix $\mathbf{T}$, as shown in Equation \ref{eq:total_sampling_matrix}.
\begin{equation}
  \mathbf{T} = [\mathbf{V}^l,\, \dots, \mathbf{V}^{j},\, \dots,\, \mathbf{V}^1]^T \in \mathbb{R}^{4l \times d}
  \label{eq:total_sampling_matrix}
\end{equation}
\vspace{-2.5em}

\begin{figure}[hbtp]
  \centering
  \begin{subfigure}[t]{0.49\linewidth}
    \centering
    \includegraphics[width=\linewidth]{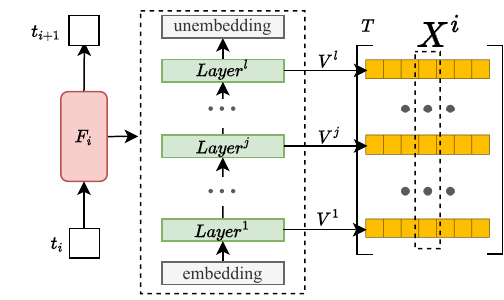}
    \caption{Construction of hidden layer temporal signals}
    \label{fig:T}
  \end{subfigure}
  \hfill
  \begin{subfigure}[t]{0.48\linewidth}
    \centering
    \includegraphics[width=\linewidth]{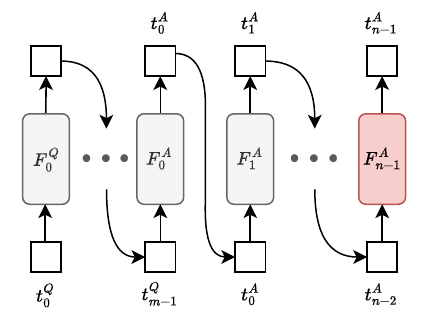}
    \caption{$F_{n-1}^A$ global reasoning implication}
    \label{fig:f}
  \end{subfigure}
  \caption{Illustration of hidden signal construction and reasoning implication}
  \label{fig:two_figs}
\end{figure}

As illustrated in Figure \ref{fig:T}, in matrix $\mathbf{T}$, each column corresponds to the temporal evolution of a specific dimension. 
For $i \in [1, d]$, the $i$-th column of matrix $\mathbf{T}$ is defined as the hidden layer temporal signal $X^i$ for the $i$-th dimension, which characterizes the variations of this dimension throughout the LLM inference process.

\subsection{LLM Hallucination Discrimination}

To determine whether the output of an LLM constitutes a hallucination, the discrimination criterion shown in Equation \ref{eq:7} is employed.

\begin{equation}
K(A \mid Q) = 
\begin{cases} 
  1, & sim(A, A^*) \leq \tau \\
  0, & otherwise
\end{cases}
\label{eq:7}
\end{equation}

Here, $Q$ represents the input question, $A$ denotes the response generated by the LLM, $A^*$ is the reference answer, $sim()$ indicates the semantic similarity score, and $\tau$ is the predetermined threshold. Samples satisfying the above condition are classified as hallucinated outputs.

\subsection{Hallucination Detection Based on Spectral Features}
Based on the constructed spectral features, we build a hallucination detector 
$H:\mathbb{R}^d \rightarrow \{0,1\}$. The input is the spectral feature $f$, 
and the output is the hallucination detection result.

For the question set $\text{Ques}(Q,A^*)$, $\forall\,Q$, the goal of the detector 
$H$ is to learn a binary predictor, as in Eq.~\ref{eq:19}.
\begin{equation}
H(Q) =
\begin{cases}
1, & if K(A\!\mid\!Q)=1 \\[2pt]
0, & otherwise
\end{cases}
\label{eq:19}
\end{equation}
Here, $K(A\!\mid\!Q)$ indicates whether the answer $A$ is hallucinated.

To balance strong nonlinear modeling capacity with deployment efficiency, 
we adopt an Enhanced MLP as the detector in the spectral-feature classification stage. 
The network consists of a feature extraction module and a classification module. 
The feature extraction module is composed of multiple fully connected layers, 
batch normalization, ReLU activation, and dropout, progressively compressing 
the feature dimensionality to $256$. The classification module is a single 
fully connected layer that maps the 256-dimensional representation to the 
binary output space $\mathbb{R}^2$.

The core computation can be described as
\begin{equation}
h = ReLU\big(BN(W f + b)\big),
\end{equation}
\begin{equation}
\hat{y} = \sigma(W h + b),
\end{equation}
where $BN()$ denotes batch normalization and $\sigma()$ is the 
Sigmoid activation function, producing the hallucination probability 
$\hat{y} \in (0,1)$. This design---staged dimensionality reduction coupled with 
regularization---captures complex spectral patterns without markedly increasing 
the parameter count and effectively suppresses overfitting.

During training, we use binary cross-entropy loss combined with $L_1$ regularization 
to control the sparsity of spectral-channel activations. The objective is given in 
Eq.~\ref{eq:22}:
\begin{equation}
  \begin{aligned}
  \mathcal{L}_{\text{halluc}} =\ & -\left[ y \log \hat{y} + (1 - y) \log (1 - \hat{y}) \right]  + \lambda \| W_1 \|_1,
  \end{aligned}
  \label{eq:22}
\end{equation}
where the first term encourages convergence to a reasonable decision boundary, 
and the $L_1$ regularizer enhances feature selectivity by attenuating 
non-discriminative frequencies, thereby improving generalization.

\section{Experiments}

\begin{table*}[tbp]
  \centering
  \begin{minipage}[t]{0.48\linewidth}
    \centering
    \caption{Comparison of LLaMA-3.1-8B and Qwen-2.5-7B-instruct model architectures}
    \label{tab:model-compare}
    \begin{tabularx}{\linewidth}{lXX}
      \toprule
      \textbf{Feature} & \textbf{LLaMA-3.1-8B} & \textbf{Qwen-2.5-7B-instruct} \\
      \midrule
      Total Parameters & 8.0B & 7.61B \\
      Number of Layers & 32  & 28  \\
      Hidden Dimension & 4096 & 3584 \\
      Context Length & 128k  & 128k \\
      Normalization & RMSNorm & RMSNorm \\
      Released by & Meta AI & Alibaba \\
      \bottomrule
    \end{tabularx}
  \end{minipage}
  \hfill
  \begin{minipage}[t]{0.48\linewidth}
    \centering
    \caption{Categories of baseline methods}
    \label{tab:baseline_methods}
    \begin{tabular}{l|l}
      \hline
      Category & Methods \\
      \hline
      \multirow{3}{*}{logit-based} & Perplexity \\
      & LN-entropy \\
      & Semantic Entropy (SE) \\
      \hline
      \multirow{3}{*}{consistency-based} & Lexical Similarity (LS) \\
      & SelfCKGPT \\
      & EigenScore \\
      \hline
      \multirow{2}{*}{verbalized} & Verbalize \\
      & Self-evaluation (Seval) \\
      \hline
      \multirow{2}{*}{Internal-state-based} & CCS \\
      & HaloScope \\
      \hline
    \end{tabular}
  \end{minipage}
\end{table*}

In this section, we first introduce the experimental setup and then present the advantages of HSAD compared with other hallucination detection methods across multiple datasets. Subsequently, we conduct detailed ablation studies to demonstrate the effects of frequency-domain modeling, cross-layer structures, key-node fusion, and the choice of appropriate hallucination observation points.

\subsection{Experimental Setup}
This subsection describes the experimental setup in terms of datasets, evaluation models, evaluation metrics, baselines, and implementation details.

\textbf{Datasets.} We evaluate on four generative question answering (QA) tasks: three open-domain QA datasets---TruthfulQA, TriviaQA, and NQ Open---as well as one domain-specific QA dataset, SciQ.  

\textbf{Model types.}   As shown in Table~\ref{tab:model-compare}, two representative models are used for evaluation: LLaMA-3.1-8B and Qwen-2.5-7B-instruct.

\textbf{Base Methods.}  As shown in Table~\ref{tab:baseline_methods}, we compare our method with a comprehensive set of baseline approaches, including state-of-the-art methods~\cite{HaloScope}. These baselines are categorized into four groups according to their underlying principles.

\subsection{Main Results}

To validate the effectiveness and generalization ability of the proposed HSAD method, hallucination detection experiments were conducted on four public datasets: TruthfulQA, TriviaQA, SciQ, and NQ Open. HSAD was systematically compared with various existing detection methods. In the experiments, the BLEURT score~\cite{bleurt2020} was used as the hallucination judgment threshold, and AUROC was adopted as the evaluation metric.

Table~\ref{tab:hallucination_prediction} reports the detection performance of different methods under various baseline models.
 It can be observed that HSAD consistently achieves the highest AUROC across all four datasets, with significant improvements over existing methods on multiple benchmarks.

\begin{table}[hbtp]
  \centering
  \caption{Hallucination detection performance of different methods on various datasets (AUROC, \%)}
  \label{tab:hallucination_prediction}
  \resizebox{\columnwidth}{!}{%
    \begin{tabular}{llcccc}
      \toprule
      \textbf{Model} & \textbf{Method} & \textbf{TruthfulQA} & \textbf{TriviaQA} & \textbf{SciQ} & \textbf{NQ Open} \\
      \midrule
      \multirow{11}{*}{Qwen-2.5-7B-instruct} 
       & Perplexity & 65.1 & 50.2 & 53.4 & 51.2 \\
       & LN-entropy & 66.7 & 51.1 & 52.4 & 54.3 \\
       & SE         & 66.1 & 58.7 & 65.9 & 65.3 \\
       & LS         & 49.0 & 63.1 & 62.2 & 61.2 \\
       & SelfCKGPT  & 61.7 & 61.3 & 58.6 & 63.4 \\
       & Verbalize  & 60.0 & 54.3 & 51.2 & 51.2 \\
       & EigenScore & 53.7 & 62.3 & 63.2 & 57.4 \\
       & Self-evaluation & 73.7 & 50.9 & 53.8 & 52.4 \\
       & CCS        & 67.9 & 53.0 & 51.9 & 51.2 \\
       & HaloScope  & 81.3 & 73.4 & 76.6 & 65.7 \\
       & \textbf{HSAD} & \textbf{82.5} & \textbf{92.1} & \textbf{94.7} & \textbf{88.3}\\
      \midrule
      \multirow{11}{*}{LLaMA-3.1-8B} 
       & Perplexity & 71.4 & 76.3 & 52.6 & 50.3 \\
       & LN-entropy & 62.5 & 55.8 & 57.6 & 52.7 \\
       & SE         & 59.4 & 68.7 & 68.2 & 60.7 \\
       & LS         & 49.1 & 71.0 & 61.0 & 60.9 \\
       & SelfCKGPT  & 57.0 & 80.2 & 67.9 & 60.0 \\
       & Verbalize  & 50.4 & 51.1 & 53.4 & 50.7 \\
       & EigenScore & 45.3 & 69.1 & 59.6 & 56.7 \\
       & Self-evaluation & 67.8 & 50.9 & 54.6 & 52.2 \\
       & CCS        & 66.4 & 60.1 & 77.1 & 62.6 \\
       & HaloScope  & 70.6 & 76.2 & 76.1 & 62.7 \\
       & \textbf{HSAD} & \textbf{81.5} & \textbf{86.7} & \textbf{85.5} & \textbf{80.7} \\
      \bottomrule
    \end{tabular}%
  }
\end{table}

In summary, HSAD demonstrates superior and stable detection performance across different datasets and baseline models, providing strong evidence of its practicality and cross-domain generalization capability for hallucination detection tasks.

\subsection{More Results}
\begin{figure}[hbtp]
  \centering
  \begin{subfigure}[t]{0.49\linewidth}
    \centering
    \includegraphics[width=\linewidth]{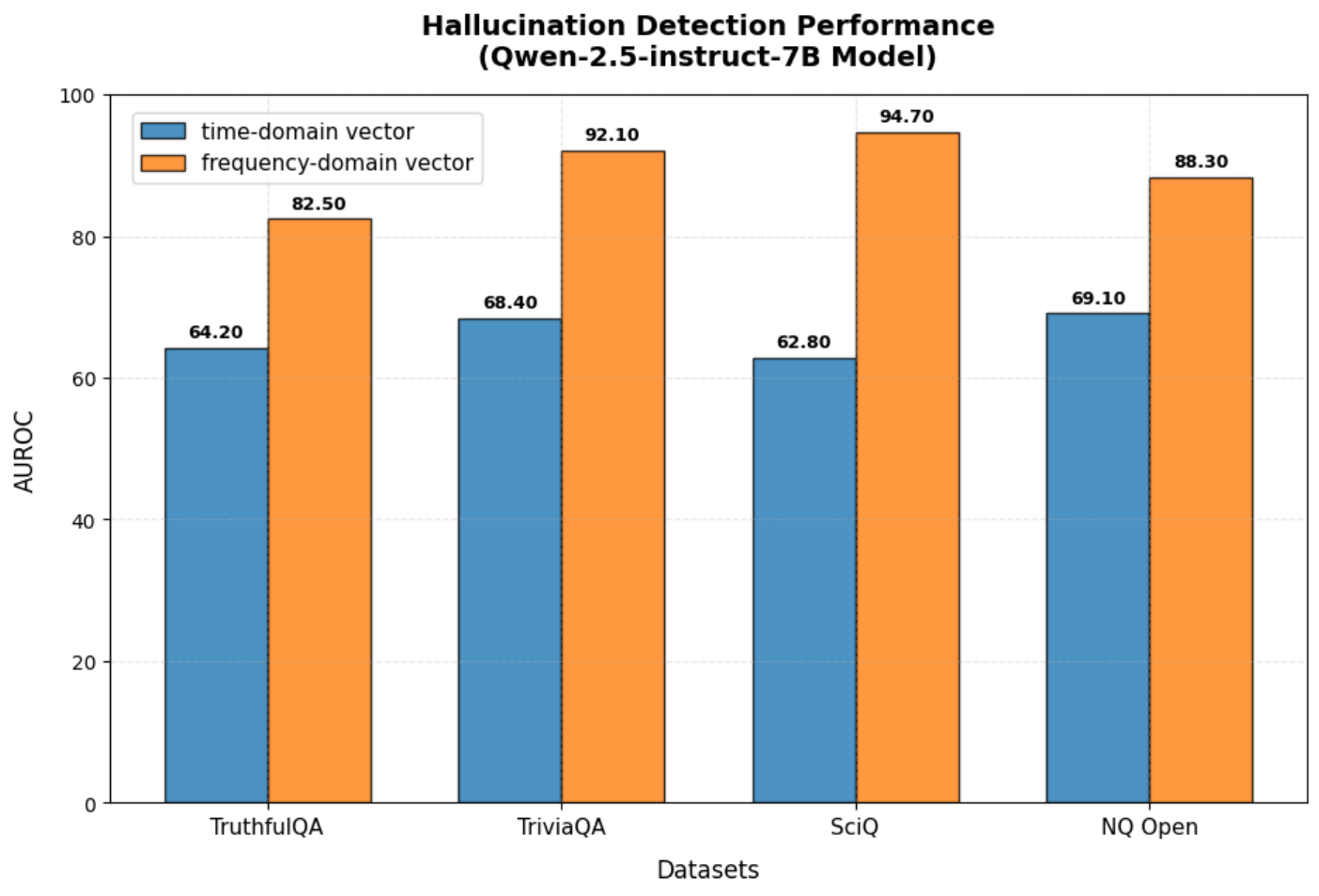}
    \caption{ Qwen}
    \label{fig:ablation_layers_qwen}
  \end{subfigure}
  \hfill
  \begin{subfigure}[t]{0.48\linewidth}
    \centering
    \includegraphics[width=\linewidth]{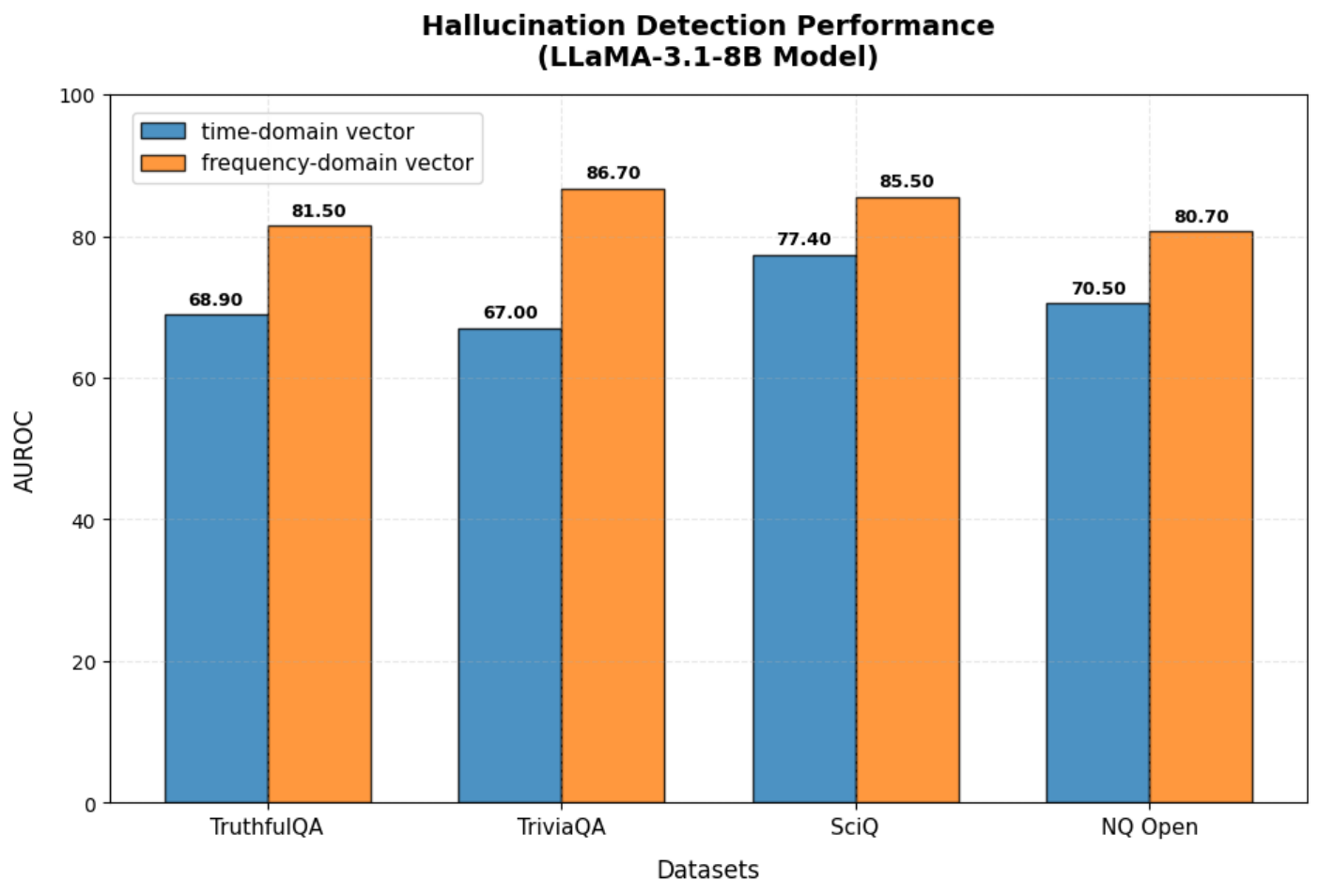}
    \caption{LLaMA}
    \label{fig:ablation_layers_llama}
  \end{subfigure}
  \caption{Effect of frequency-domain modeling on detection performance}
  \label{fig:two_figs}
\end{figure}

\textbf{Effectiveness of Frequency-domain Modeling.} To test whether spectral features provide essential discriminative information for hallucination detection, we replaced the Fast Fourier Transform  module in the original HSAD with a direct use of hidden state time-series signals. In this variant, the maximum value of each hidden dimension was taken as the representation, while keeping all other settings identical. 
As shown in Fig.~\ref{fig:ablation_layers_qwen} and Fig.~\ref{fig:ablation_layers_llama}, the full cross-layer modeling strategy significantly outperforms the other variants. This indicates that inter-layer dynamics encode essential reasoning features, which are critical for hallucination identification.

\begin{figure}[hbtp]
  \centering
  \begin{subfigure}[t]{0.49\linewidth}
    \centering
    \includegraphics[width=\linewidth]{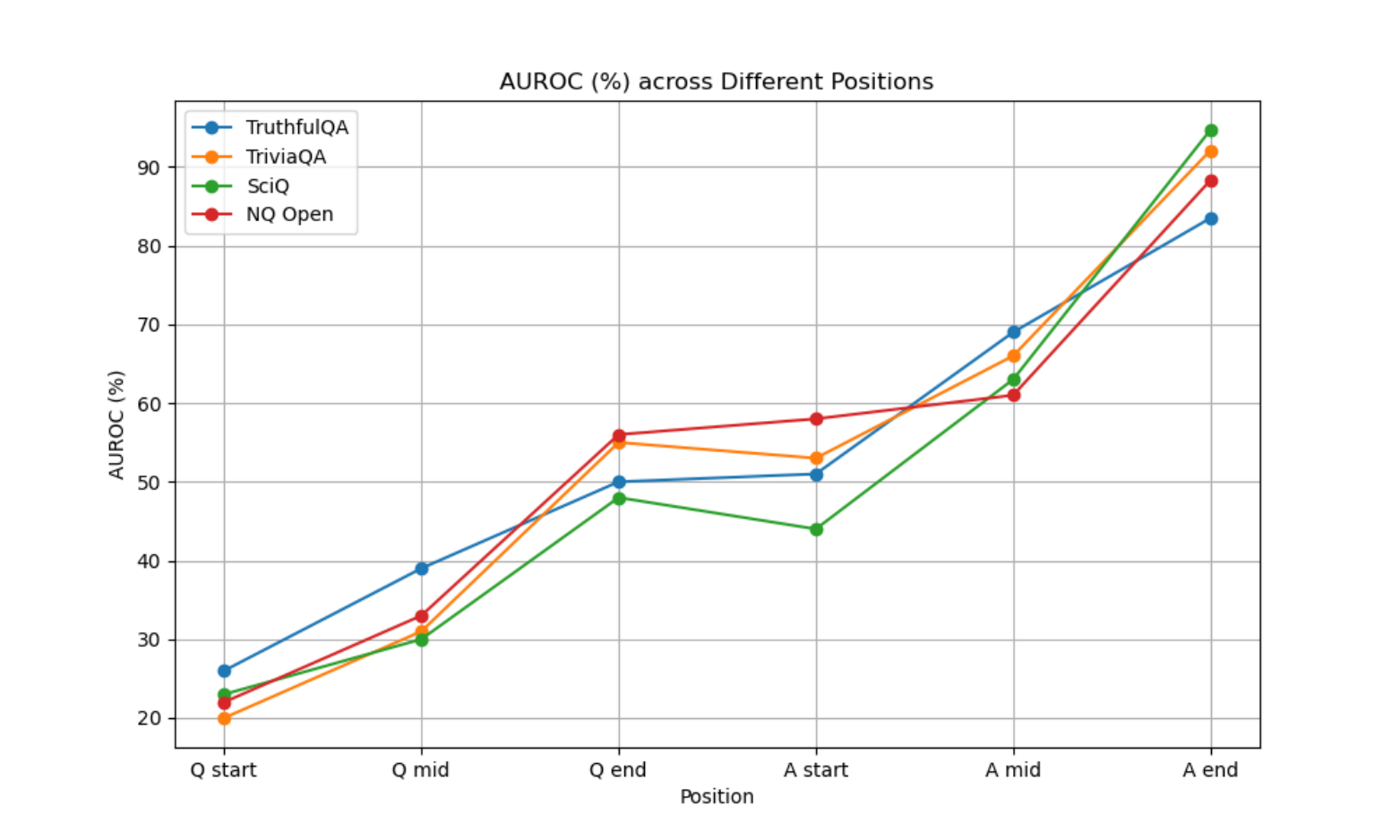}
    \caption{ Qwen}
    \label{fig:positions_qwen}
  \end{subfigure}
  \hfill
  \begin{subfigure}[t]{0.48\linewidth}
    \centering
    \includegraphics[width=\linewidth]{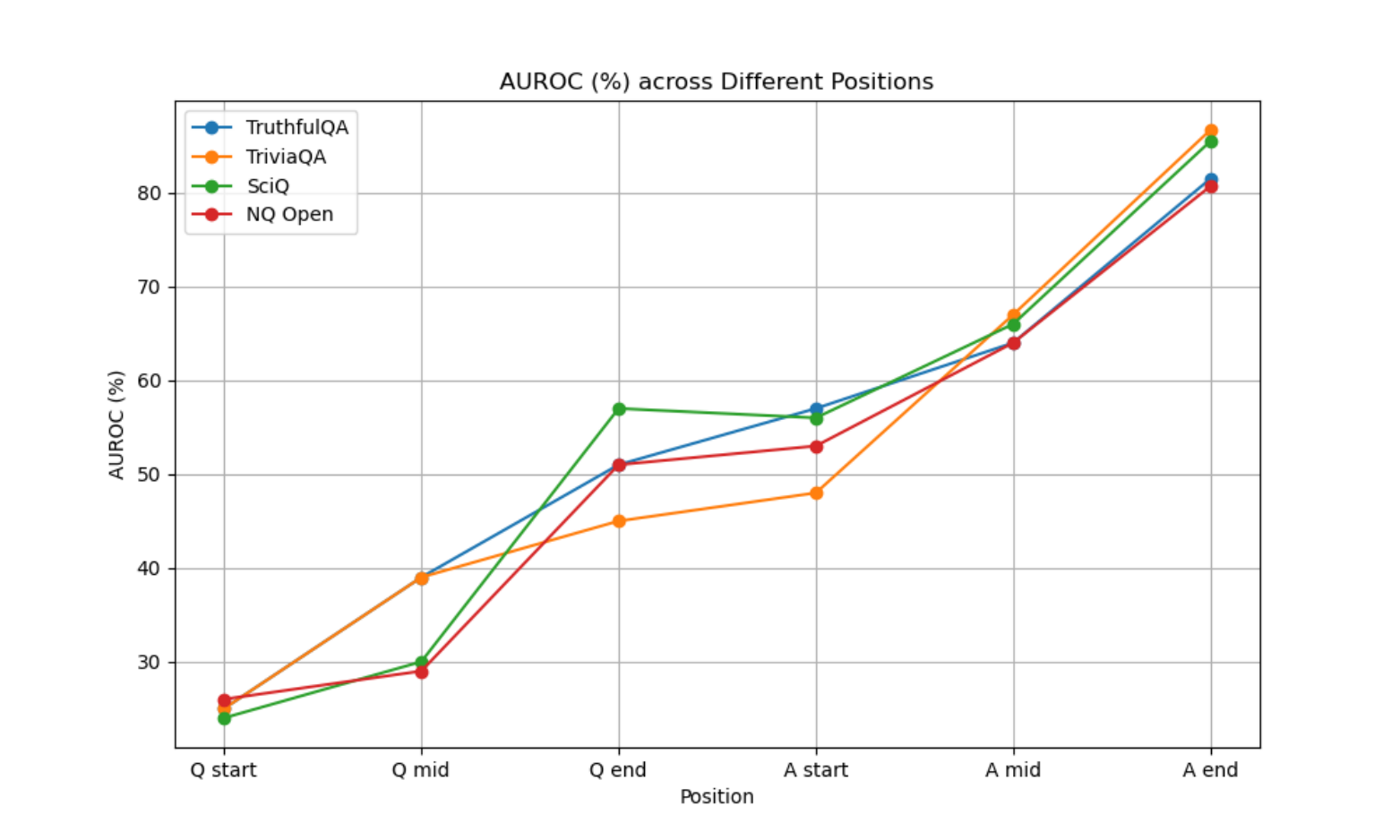}
    \caption{LLaMA}
    \label{fig:positions_llama}
  \end{subfigure}
  \caption{Comparison of observation point positions}
  \label{fig:two_figs}
\end{figure}

\textbf{Analysis of Observation Point Positions.} To assess the influence of different observation points on hallucination detection, we evaluated several positions: Q start, Q mid, Q end, A start, A mid, and A end. 
As illustrated in Fig.~\ref{fig:positions_qwen} and Fig.~\ref{fig:positions_llama}, different observation points lead to significant differences in detection performance. In particular, A end (the end of the generated answer) achieves the best results, outperforming positions in the question segment (Q start/mid/end) and the earlier answer positions (A start, A mid). 

\begin{figure}[hbtp]
  \centering
  \begin{subfigure}[t]{0.49\linewidth}
    \centering
    \includegraphics[width=\linewidth]{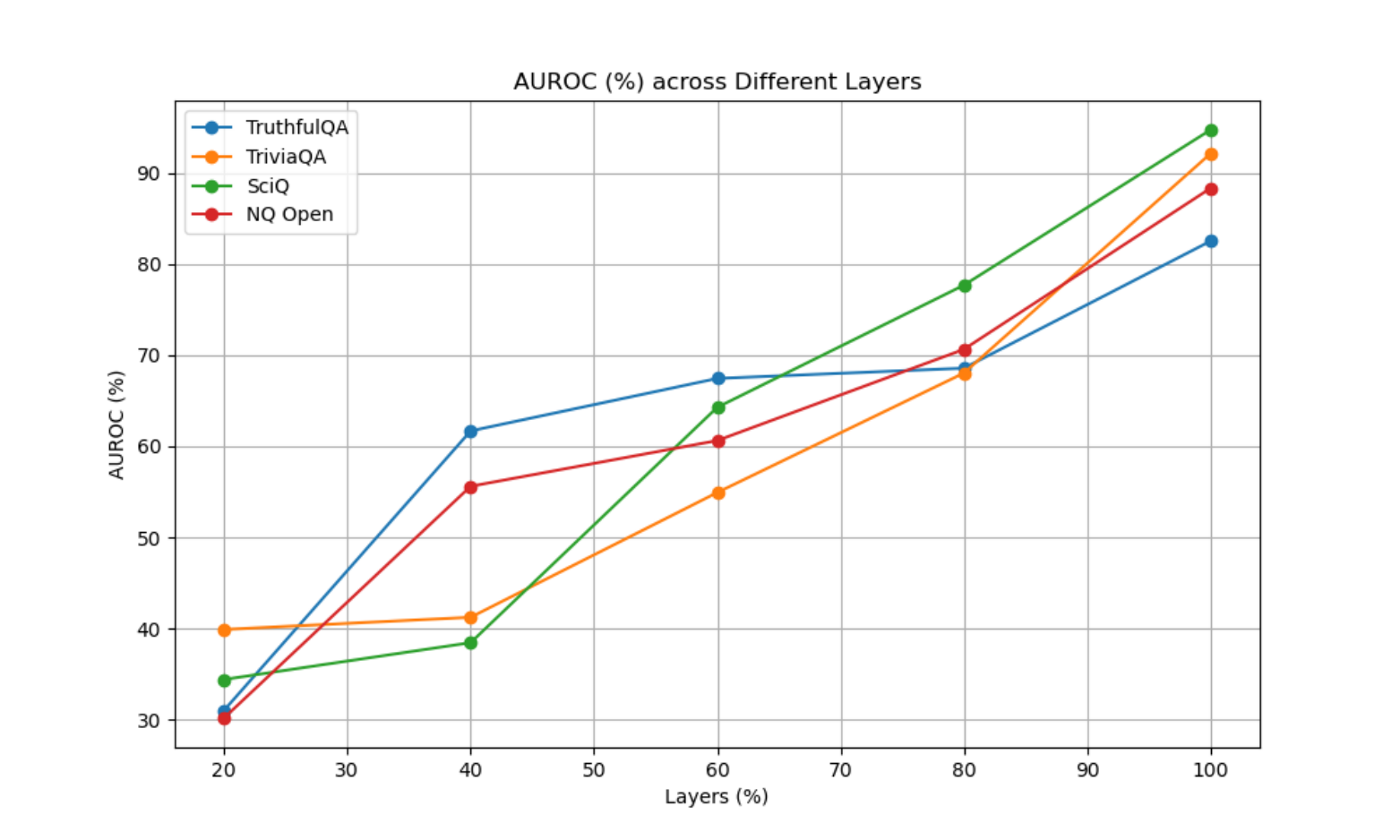}
    \caption{ Qwen}
    \label{fig:layers_qwen}
  \end{subfigure}
  \hfill
  \begin{subfigure}[t]{0.48\linewidth}
    \centering
    \includegraphics[width=\linewidth]{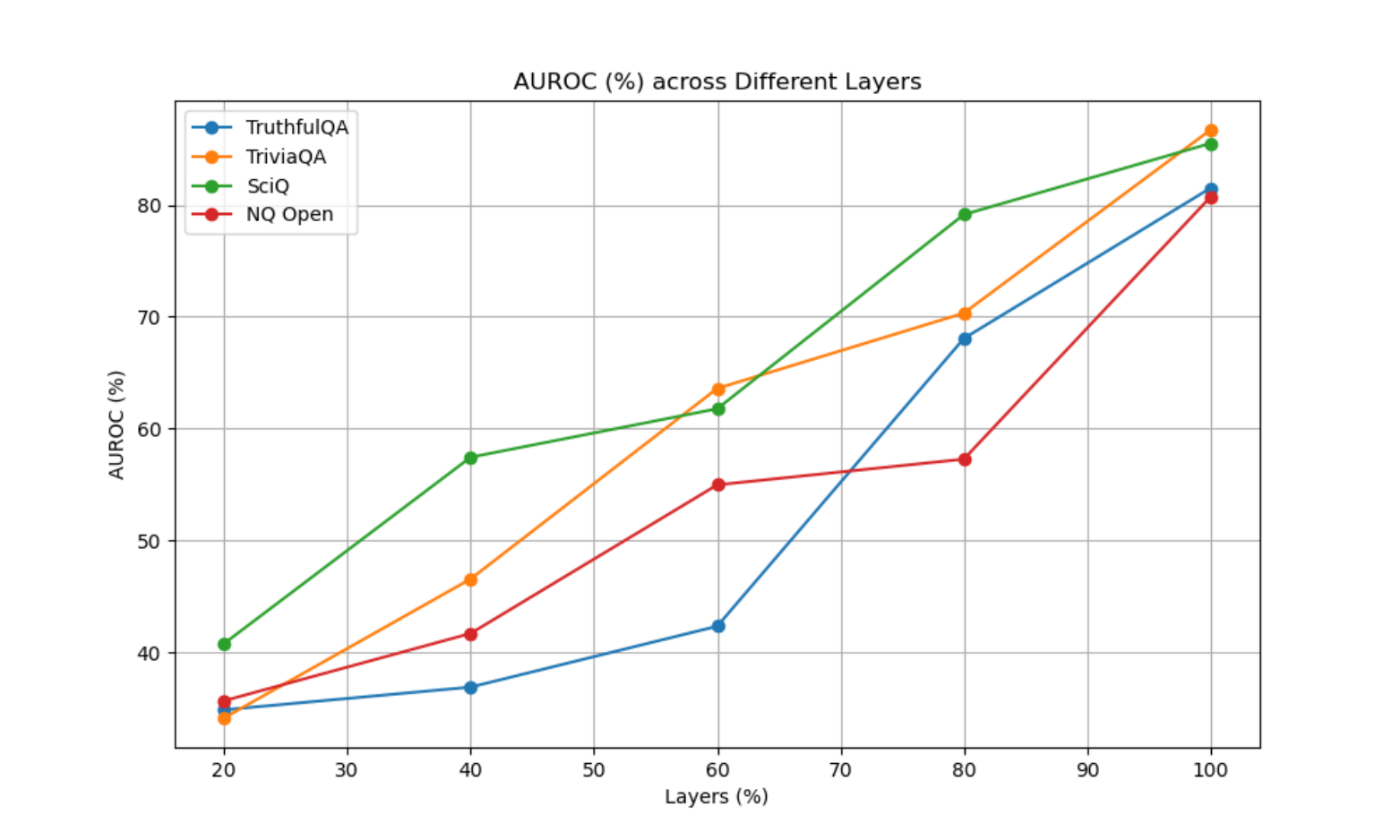}
    \caption{LLaMA}
    \label{fig:layers_llama}
  \end{subfigure}
  \caption{Effect of random layer sampling on performance}
  \label{fig:two_figs}
\end{figure}

\textbf{Cross-layer Structure Analysis.} To evaluate the impact of cross-layer modeling on LLM hallucination detection,We  compared HSAD’s performance under different sampling rates across datasets. 
Fig.~\ref{fig:layers_qwen} and Fig.~\ref{fig:layers_llama} illustrate the relationship between the number of sampled layers and detection performance. As the number of sampled layers increases, performance steadily improves and saturates when all layers are included. This further verifies the integrative nature of spectral features along the layer dimension, highlighting the importance of treating the reasoning process as a holistic temporal cognitive system for modeling.

\section{Conclusion}
This study proposes a hallucination detection method for LLMs inspired by human polygraph mechanisms—HSAD (\textbf{H}idden \textbf{S}ignal \textbf{A}nalysis-based \textbf{D}etection). 
The method constructs hidden-layer time-series signals from the reasoning process of LLMs and applies the Fast Fourier Transform for frequency-domain modeling, thereby capturing potential hallucinations during generation. 
By extracting spectral features, HSAD can effectively identify abnormal signals related to LLM hallucinations without relying on external knowledge, overcoming the limitations of existing approaches that depend heavily on knowledge coverage or struggle to capture deviations in the reasoning process. 
This study not only provides a novel perspective for analyzing the mechanisms of LLM hallucinations but also offers a theoretical foundation and practical pathway for implementing safety control mechanisms in high-reliability language generation tasks.

%Bibliography
\bibliographystyle{unsrt}  
\bibliography{references}

\end{document}